\documentclass[letterpaper, 10 pt, conference]{ieeeconf}  % Comment this line out if you need a4paper

\IEEEoverridecommandlockouts                              % This command is only needed if 
\usepackage{graphics} % for pdf, bitmapped graphics files
\usepackage{epsfig} % for postscript graphics files
\usepackage{amsmath} % assumes amsmath package installed
\usepackage{amssymb}  % assumes amsmath package installed
\usepackage{dsfont}
\usepackage{hyperref}
\usepackage{multirow}
\usepackage{mathrsfs}
\usepackage{commath}
\usepackage{flexisym}
\usepackage{booktabs,caption}
\usepackage[flushleft]{threeparttable}
\usepackage{float}

\title{\LARGE \bf
A LiDAR Point Cloud Generator: from a Virtual World to \\Autonomous Driving
}

\author{Xiangyu Yue, Bichen Wu, Sanjit A. Seshia, Kurt Keutzer and Alberto L. Sangiovanni-Vincentelli\\
University of California, Berkeley\\
\{xyyue, bichen, sseshia, keutzer, alberto\}@eecs.berkeley.edu
}

\begin{document}

\maketitle
\thispagestyle{empty}
\pagestyle{empty}

%%%%%%%%%%%%%%%%%%%%%%%%%%%%%%%%%%%%%%%%%%%%%%%%%%%%%%%%%%%%%%%%%%%%%%%%%%%%%%%%
\begin{abstract}
%State of the art algorithms in computer vision and robotics for autonomous driving have mostly been based on high-capacity deep neural networks trained on large datasets. 
3D LiDAR scanners are playing an increasingly important role in autonomous driving as they can generate depth information of the environment. However, creating large 3D LiDAR point cloud datasets with point-level labels requires a significant amount of manual annotation. This jeopardizes the efficient development of supervised deep learning algorithms which are often data-hungry. 
We present a framework to rapidly create point clouds with accurate point-level labels from a computer game. The framework supports data collection from both auto-driving scenes and user-configured scenes. Point clouds from auto-driving scenes can be used as training data for deep learning algorithms, while point clouds from user-configured scenes can be used to systematically test the vulnerability of a neural network, and use the falsifying examples to make the neural network more robust through retraining. In addition, the scene images can be captured simultaneously in order for sensor fusion tasks, with a method proposed to do automatic calibration between the point clouds and captured scene images. We show a significant improvement in accuracy (+9\%) in point cloud segmentation by augmenting the training dataset with the generated synthesized data. 
%Point clouds from user-configured scenes can be used to systematically test, analyze and improve performance of neural networks. 
Our experiments also show by testing and retraining the network using point clouds from user-configured scenes, the weakness/blind spots of the neural network can be fixed. 
 \end{abstract}

%%%%%%%%%%%%%%%%%%%%%%%%%%%%%%%%%%%%%%%%%%%%%%%%%%%%%%%%%%%%%%%%%%%%%%%%%%%%%%%%
 \section{Introduction}
%State of the art algorithms in computer vision and robotics for autonomous driving have mostly been based on high-capacity deep neural networks trained on large datasets. 
Autonomous driving requires accurate and reliable perception of the environment. 
Of all the environment sensors, 3D LiDARs (Light Detection And Ranging) play an increasingly important role, since their resolution and field of view exceed radar and ultrasonic sensors and they can provide direct distance measurements that allow detection of all kinds of obstacles \cite{3dlidar}. Moreover, LiDAR scanners are robust under a variety of conditions: day or night, with or without glare and shadows \cite{squeezeSeg}. While LiDAR point clouds contain accurate depth measurement of the environment, navigation of autonomous vehicles also relies on correct understanding of the semantics of the environment. Most of the LiDAR-based perception tasks, such as semantic segmentation\cite{segmentation1, segmentation2, squeezeSeg} and drivable area detection\cite{drivable1, drivable2}, require significant amount of point-level labels for training and/or validation. Such annotation, however, is usually very expensive.

%which could provide further information for the vehicle controllers and planners. 

To facilitate the manual annotation process, much work has been done on interactive annotation. Annotation methods have been proposed for labeling 3D point clouds of both indoor scenes \cite{indoorRGBD} and outdoor driving scenes \cite{KITTI}. These methods utilize little computer assistance during the annotation process and thus need a significant amount of human effort. In \cite{rapid3DSelection,goThenTag}, approaches have been proposed to enhance the man-machine interaction  to improve annotation efficiency. In \cite{smartAnnotator,interactiveRGBD}, annotation suggestions for indoor RGBD scenes are proposed by the system that are interactively corrected or refined by the user. In order to provide faster interactive labeling rates, \cite{groupAnnotation} proposes a group annotation approach for labeling objects in 3D LiDAR scans. 
Active learning has also been introduced in the annotation process to train a classifier with fewer interactions \cite{activelearning1,activelearning2}, yet it requires users to interact with examples one-by-one. Other frameworks further take into account the risk of mislabeling and cost of annotation. \cite{onlineCrowdsourcing} proposes a model of the labeling process and dynamically chooses which images will be labeled next in order to achieve a desired level of confidence. 

\begin{figure*}
  \centering
  \includegraphics[width=6.8in, trim={0cm 1cm 0cm 0cm}]{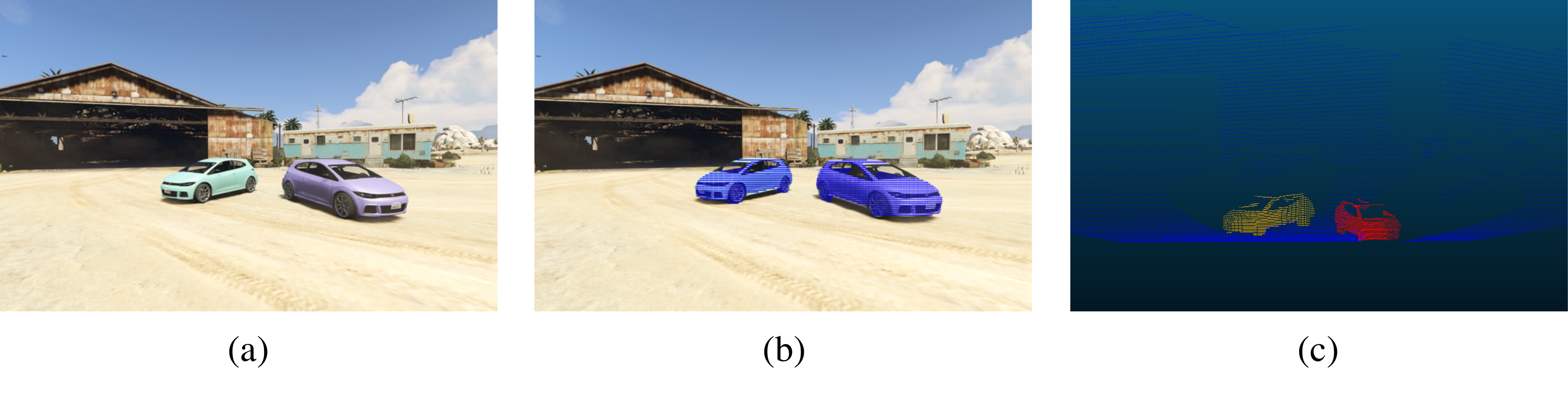}
  %\fbox{\rule[-.5cm]{0cm}{4cm} \rule[-.5cm]{4cm}{0cm}}
  \caption{Sample data extracted from an in-game scene. (a): Image of the scene; (b): Point cloud of car (Blue dots) mapped to image after calibration matches car in image; (c): Extracted point cloud from the same scene.}
  \label{fig:scene}
\end{figure*}

Recently, video games have been used for creating large-scale ground truth data for training purposes. In \cite{playingForData}, a video game is used to generate ground truth semantic segmentation for the synthesized in-game images. However, human effort is still required in the annotation process. In \cite{drivingInMatrix}, the same game engine is used to generate ground truth 2D bounding boxes of objects in the images. \cite{playing_for_benchmarks} further extends the work of \cite{playingForData} so that various ground truth information(e.g. semantic segmentation, semantic
instance segmentation, and optical flow) can be extracted from the game engine. In addition, many driving simulation environments\cite{congrats, simulator, carla} have been built in order to obtain various kinds of labeled data for autonomous driving purposes. Many of these work\cite{drivingInMatrix, playingForData, playing_for_benchmarks, carla} show the effectiveness of synthetic data in image-based learning tasks by showing improved performance after training with additional synthetic data. 
However, little work has been done on extracting annotated 3D LiDAR point clouds from simulators, not to mention showing the efficacy of the synthetic point clouds during the training process of neural networks.

Note that even if we could provide large amounts of training data, it is still almost impossible for any algorithms to achieve 100\% accuracy. For Cyber-Physical Systems used for safety critical purposes, such as autonomous driving, verifying neural networks is of  extreme importance \cite{dreossiDS17}. In \cite{falsifyCNN}, a framework is proposed to systematically analyze Convolutional Neural Networks (CNNs) used in objection detection in autonomous driving systems. However, the framework only takes into account cars from direct front/back view and thus has a very limited modification space. In addition, each background image needs to be manually annotated, making it expensive to generate a dataset with large diversity. To the best of our knowledge, no similar work has been done on LiDAR point clouds. 
In this paper, we propose an extraction-annotation-CNN testing framework based on a popular video game. The main contributions of this framework are as follows: 

\begin{itemize}
	\item The framework can automatically extract point-cloud data with ground truth labels together with the corresponding image frame of the in-game scene, as shown in Fig. \ref{fig:scene}.
	\item The framework can do automatic calibration between collected point clouds and images which can then be used together for sensor fusion tasks.
	\item Users can construct specified scenarios in the framework interactively and the collected data(point clouds and images) can then be used to systematically test, analyze and improve LiDAR-based and/or image-based learning algorithms for autonomous driving. 
\end{itemize}

We conducted experiments on a Convolutional Neural Network (CNN)-based model for 3D LiDAR point cloud segmentation using the data collected from the proposed framework. The experiments show 1) significantly improved performance on KITTI dataset\cite{KITTI} after retraining with additional synthetic LiDAR point clouds, and 2) efficacy of using the data collected from user-configured scenes in the framework to test, analyze and improve the performance of the neural network. The performance improvements come from the fact that the data collected in the rich virtual world contains a lot of information that the neural network failed to learn from the limited amount of original training samples. 

% In this paper, we propose a framework based on a popular video game that can address both issues of labeled data generation and systematic testing of neural nets. First, our framework can automatically extract point cloud data with ground truth labels together with the corresponding image frame of the in-game auto-driving scene, as shown in Figure \ref{fig:scene}. The collected point cloud dataset itself can then be used as training data, and the corresponding images can be further used for sensor fusion algorithms. Second, in our framework, the user can configure desired scenes interactively. The collected data can then be used for neural network testing and analysis. Preliminary experimental results show significant advantages over state-of-the-art methods.

\section{Technical Approach}
\label{sec:technical_approach}
\subsection{In-Game Simulation Setup and Method for Data Collection} \label{ssec:setup}

We choose to utilize the rich virtual world in Grand Theft Auto V (GTA-V), a popular video game, to obtain simulated point clouds as well as captured in-game images with high fidelity\footnote[1]{The publisher of GTA-V allows non-commercial use of footage of gameplay \cite{playingForData}.}.  Our framework is based on DeepGTAV\footnote[2]{\url{https://github.com/aitorzip/DeepGTAV}}, which uses Script~Hook~V\footnote[3]{\url{http://www.dev-c.com/gtav/scripthookv/}} as a plugin. 

In order to simulate realistic driving scenes, an ego car is used in the game with a virtual LiDAR scanner mounted atop, and it is set to drive autonomously in the virtual world with the AI interface provided in Script Hook V. While the car drives on a street, the system collects LiDAR point clouds and captures the game screen, simultaneously. We place the virtual LiDAR scanner and the game camera at the same position in the virtual 3D space. This set-up offers two advantages: 1) a sanity check can be easily done on the collected data, since point clouds and corresponding images must be consistent; 2) calibration between the game camera and the virtual LiDAR scanner can be done automatically, and then collected point clouds and scene images can be combined together as training dataset for neural networks for sensor fusion tasks. Details of the proposed calibration method will be described in Section \ref{ssec:calibration}. 

Ray casting is used to simulate each laser ray emitted by the virtual LiDAR scanner. The ray casting API takes as input the 3D coordinates of the starting and ending point of the ray, and returns the 3D coordinates of the first point the ray hits. This point is used, with another series of API function calls, to calculate, among other data, the distance of the point, the category and instance ID of the object hit by the ray, thus allowing automatic annotation on the collected data. 

\begin{figure}[t]
  \centering
  \vspace{0.5cm}
  \includegraphics[width=3.3in, trim={0.3cm 0.9cm 0cm 1.3cm}]{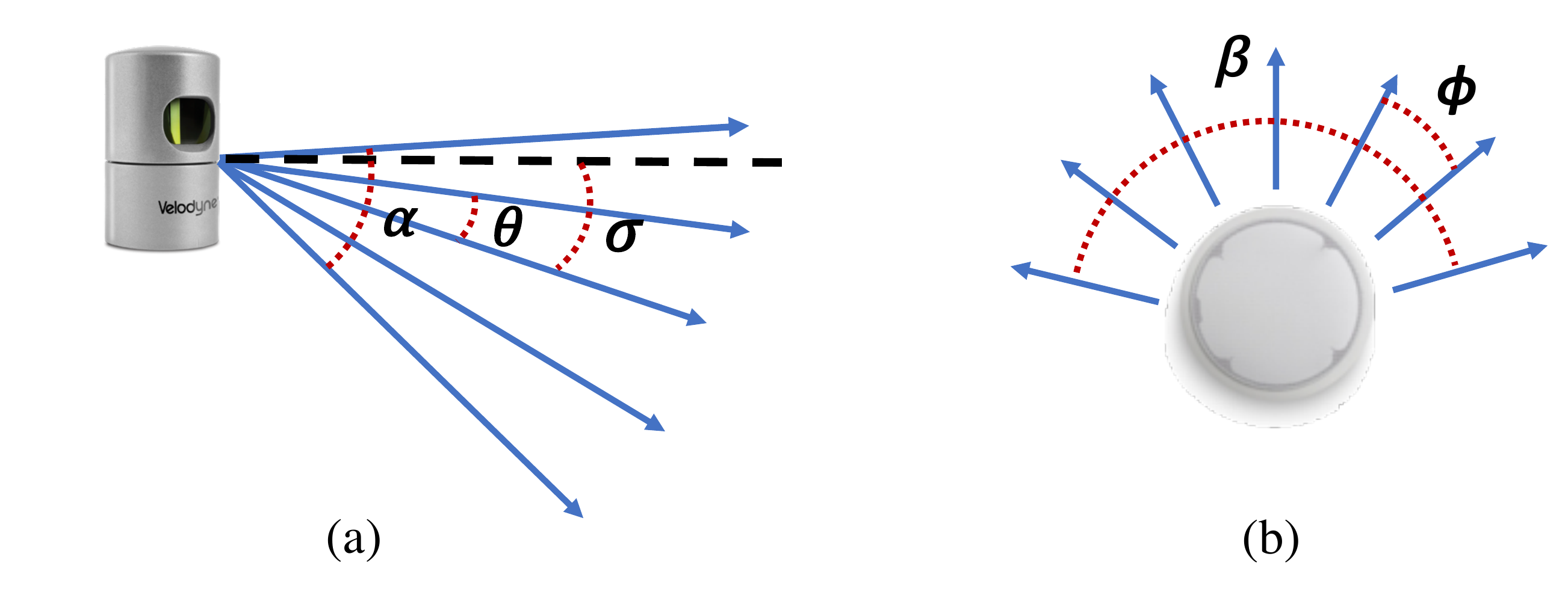}
  %\fbox{\rule[-.5cm]{0cm}{4cm} \rule[-.5cm]{4cm}{0cm}}
  \caption{Sample configurable parameters of the virtual LiDAR. (a) shows front view of the virtual LiDAR: black dotted line is the horizontal line, $\alpha$ is the vertical field of view (FOV), $\theta$ is the vertical resolution, $\sigma$ is the pitch angle; (b) shows top view of the virtual LiDAR, $\beta$ is the horizontal FOV, and $\phi$ is the horizontal resolution.}
  \label{fig:LiDARScanner}
\end{figure}

In our framework, users can provide configurations of the LiDAR scanner including vertical field of view (FOV), vertical resolution, horizontal FOV, horizontal resolution, pitch angle, maximum range of laser rays, and scanning frequency. Some of the configurable parameters are shown in Fig. \ref{fig:LiDARScanner}.

% \begin{figure*}[h]
%     \centering
%     \includegraphics[width=7in]{figs/project.pdf}
%     \caption{LiDAR Projections. Note that each channel reflects structural information in the camera-view image.}
%     \label{fig:ProjectedData}
% \end{figure*}
\subsection{Automatic Calibration Method} \label{ssec:calibration}
\begin{figure}
  \centering
  \includegraphics[width=3.3in, trim={1.4cm 6cm 0cm 7.5cm}]{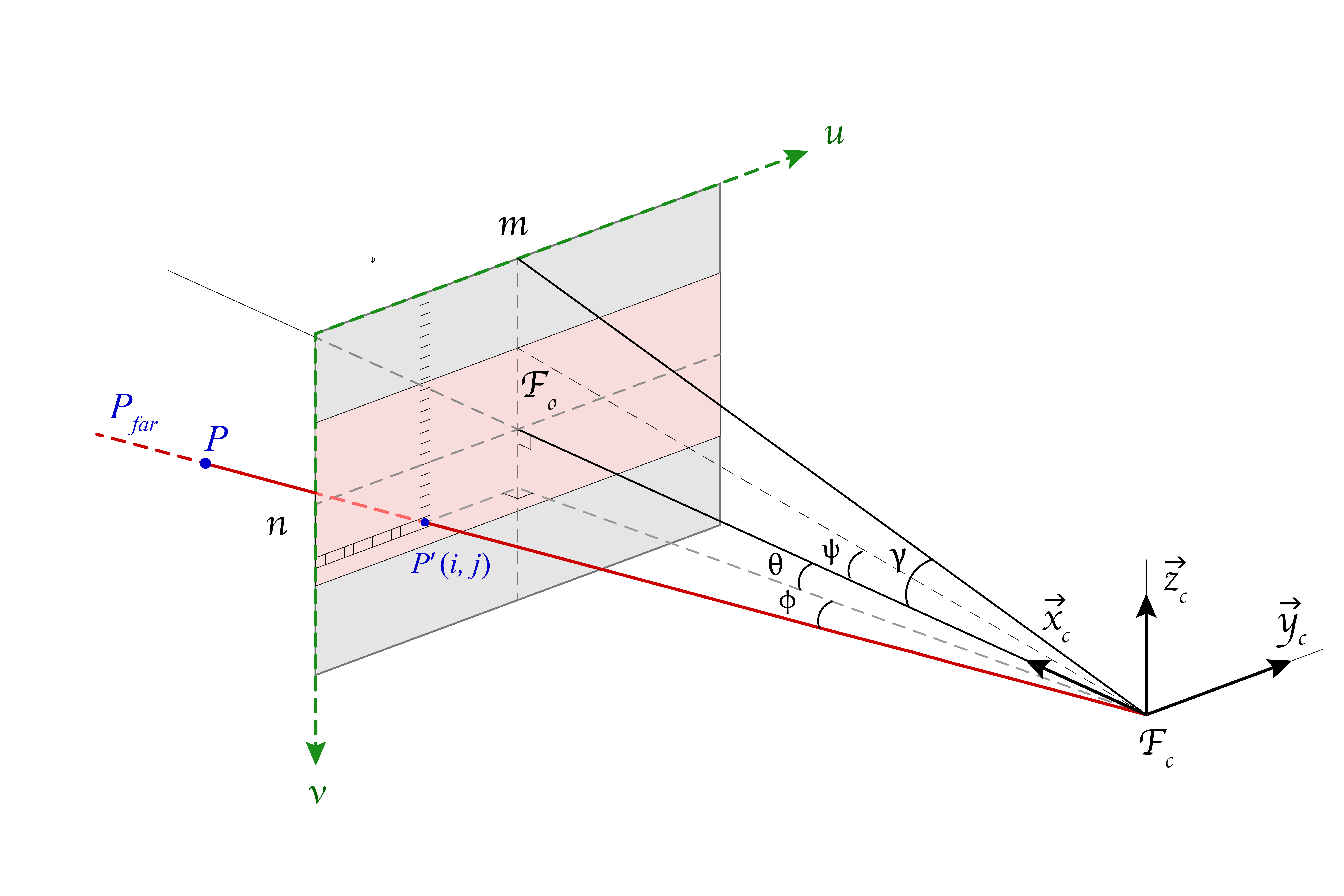}
  %\fbox{\rule[-.5cm]{0cm}{4cm} \rule[-.5cm]{4cm}{0cm}}
  \caption{Projection for Calibration. $\mathcal{F}_o$ is the center of the near clipping plane of the camera; $\mathcal{F}_c$ is the center of the camera and of the LiDAR scanner; the red line is the laser ray and \textit{P} is the point hit by the ray; the calibrated on-image point has pixel index $(\textit{i}, \textit{j})$ and 3D coordinates $P\textprime$; $\gamma$ is the $1/2$ camera vertical FOV and $\psi$ is the $1/2$ LiDAR vertical FOV; $\phi$ and $\theta$ are the $azimath$ and $zenith$ angles of the laser ray.}
  \label{fig:calibration}
\end{figure}

The goal of the calibration process is to find the corresponding pixel in the image for each LiDAR point. In our framework, the calibration process can be done automatically by the system based on the parameters of the camera and LiDAR scanner. In addition, the centers of the camera and LiDAR scanner are set to the same position in the virtual world, making the calibration projection similar to the camera perspective projection model, as shown in Fig. \ref{fig:calibration}.

The problem is formulated as follows: for a certain laser ray with $azimuth$ angle $\phi$ and $zenith$ angle $\theta$, calculate the index $(\textit{i}, \textit{j})$ of the corresponding pixel on image. 
$\mathcal{F}_c$, $\mathcal{F}_o$, $P$, $P\textprime$ and $P_{far}$ are 3D coordinates of a) center of camera/LiDAR scanner, b) center of camera near clipping plane. c) point first hit by the virtual laser ray (in red), d) pixel on image corresponding to $P$, and e) a point far away in the laser direction, respectively.
$m$ and $n$ are the width and height of the near clipping plane. $\gamma$ is $1/2$ vertical FOV of camera while $\psi$ is $1/2$ vertical FOV of the LiDAR scanner. Note that LiDAR scanner FOV is usually smaller than camera FOV, since there is usually no object in
the top part of the image, and the emitting laser to open space is not necessary.
After a series of 3D geometry calculation, we can get:
\begin{gather}
%\label{eqn:projection}
\begin{split}
% \bar{u} &= f \cdot \tan \gamma\cdot\dfrac{m}{n} - \dfrac{f}{\cos\theta}\cdot\tan\phi, \\
% \bar{v} &= f \cdot \tan \gamma + f \cdot \tan\theta, \\
i &= \dfrac{R_m}{m} \cdot (f \cdot \tan \gamma\cdot\dfrac{m}{n} - \dfrac{f}{\cos\theta}\cdot\tan\phi), \\
j &= \dfrac{R_n}{n} \cdot (f \cdot \tan \gamma + f \cdot \tan\theta),
\end{split}
\end{gather}
where $f = \norm{\overrightarrow{\mathcal{F}_c\mathcal{F}_o}}$, and $(R_m, R_n)$ is the pixel resolution of the image/near clipping plane.

Further, in order for the ray casting API to work properly, the 3D coordinates of $P_{far}$ are also required. Using similar 3D geometry calculations, we obtain:
\begin{gather}
\begin{split}
&P\textprime = \mathcal{F}_c + f\cdot \overrightarrow{x_c} - \dfrac{f}{\cos\theta}\cdot\tan\phi \cdot \overrightarrow{y_c} - f \cdot \tan\theta\cdot \overrightarrow{z_c}, \\
&P_{far} = \mathcal{F}_c + k \cdot(P\textprime - \mathcal{F}_c),
\end{split}
\end{gather}
where $k$ is a large coefficient, and $\overrightarrow{x_c}, \overrightarrow{y_c}, \overrightarrow{z_c}$ are unit vectors of the camera axis in the world coordinate system. 
% $$
% \bar{u} = f \cdot \tan \gamma\cdot\dfrac{m}{n} - \dfrac{f}{\cos\theta}\cdot\tan\varphi,
% $$
% $$
% \bar{v} = f \cdot \tan \gamma + f \cdot \tan\theta,
% $$

An example of the calibration result is shown in Fig. \ref{fig:scene}. After simulation, both image and point cloud of the specified in-game scene are collected by the framework (Fig. \ref{fig:scene} (a, c)). Then with the proposed calibration method, we map all the points with category "Car" to the corresponding image. As shown in Fig. \ref{fig:scene} (b),  the mapped car point cloud (blue dots) matches the car in the image fairly accurately. 

% width=3.5in,trim={1cm 6cm 0cm 5.5cm},clip
\vspace{-1.5 mm}
\subsection{Configurable In-game Scene}

\begin{figure*}[h]
  \centering
  \includegraphics[width=5.5in, trim={0cm 1cm 0cm 0cm}]{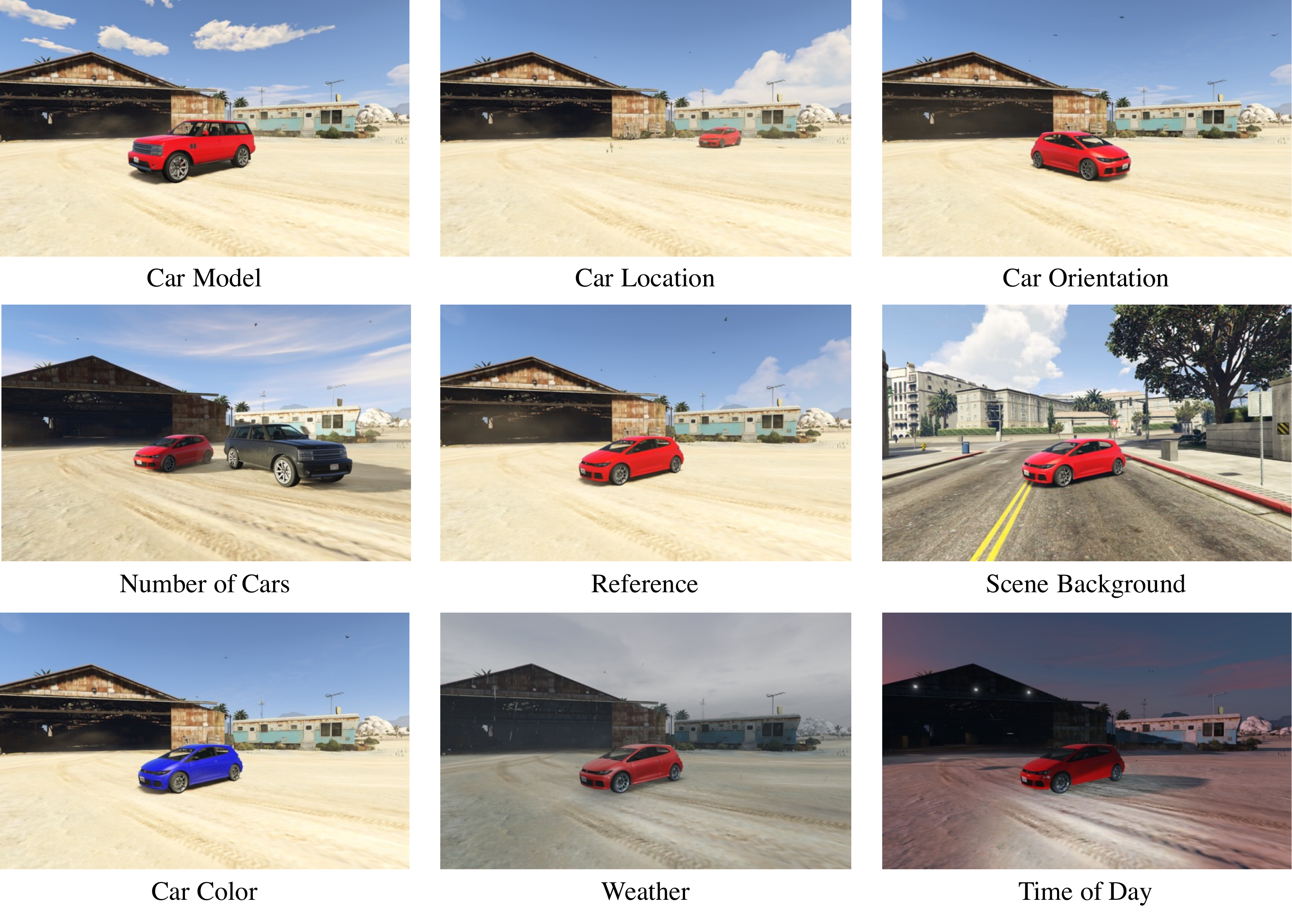}
  %\fbox{\rule[-.5cm]{0cm}{4cm} \rule[-.5cm]{4cm}{0cm}}
  \caption{Modification dimensions of the framework with image in center showing the reference scene.}
  \label{fig:dimensions}
\end{figure*}

\begin{figure*}[t]
  \centering
  \includegraphics[width=5.5in, trim={0.5cm 1cm 0cm 0cm}]{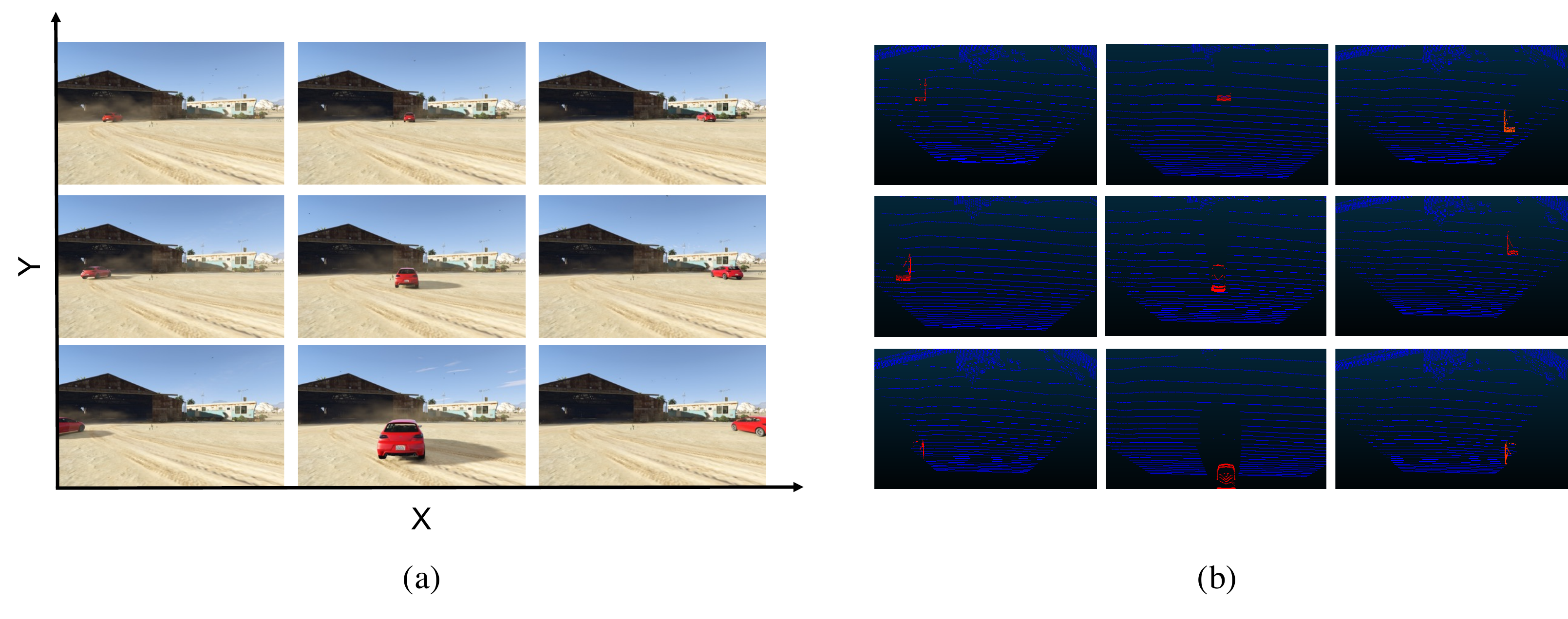}
  %\fbox{\rule[-.5cm]{0cm}{4cm} \rule[-.5cm]{4cm}{0cm}}
  \caption{Scenes with one car sampled from spatial dimensions  and corresponding point cloud. (a) shows the scene image while changing the location of the car on X(left-right) and Y(forward-backward) directions; (b) shows point clouds (red for car and blue for background) of scenes in (a). }
  \label{fig:exp_with_lidar}
\end{figure*}

Besides the auto-driving mode for large-scale data collection, our framework offers a configurable mode, where the user can configure desired in-game scenes and collect data from them. 
One advantage of configurable scenes is generating training data of driving scenes that are dangerous or rare in real world. Another advantage is that we can systematically sample the modification space(e.g. number of cars, position and orientation of a car) of an in-game scene. The data can then be used to test neural network, expose its vulnerabilities and improve its performance throuth retraining. 
Our framework offers a large modification space of the in-game scene. As shown in Fig. \ref{fig:dimensions}, the user can specify and change 8 dimensions of in-game scene: car model, car location, car orientation, number of cars, scene background, color of car, weather, and time of day. The first 5 dimensions affect both LiDAR point cloud and scene image, while the last three dimensions affect only the scene image. An example of sampling is shown in Fig. \ref{fig:exp_with_lidar}, where the scenes are only sampled from the spatial dimensions (X, Y) with only one car in each scene. X and Y are the location offset of the car relative to the camera/LiDAR location in the left-right and forward-backward directions. Fig.\ref{fig:exp_with_lidar} (b) shows collected point cloud of the samples shown in Fig.\ref{fig:exp_with_lidar} (a). The red points represent car points while the blue points represent the scene background. The collected point clouds match the scenes well thus allowing the use of the data to test neural nets systematically. %, which is further discussed in the next Section. 

\section{Experiments and Results}
\label{sec:experiments}

We performed experiments to show the efficacy of our data synthesis framework: 1) Data collected by the framework can be used in the training phase and help improve the validation accuracy; 2) Collected data can be used to systematically test a neural network and improve its performance via retraining. 
\subsection{Evaluation Metrics}
\begin{figure}[h!]
  \centering
  \includegraphics[width=3.4in, trim={0cm 0cm 0cm 0cm}]{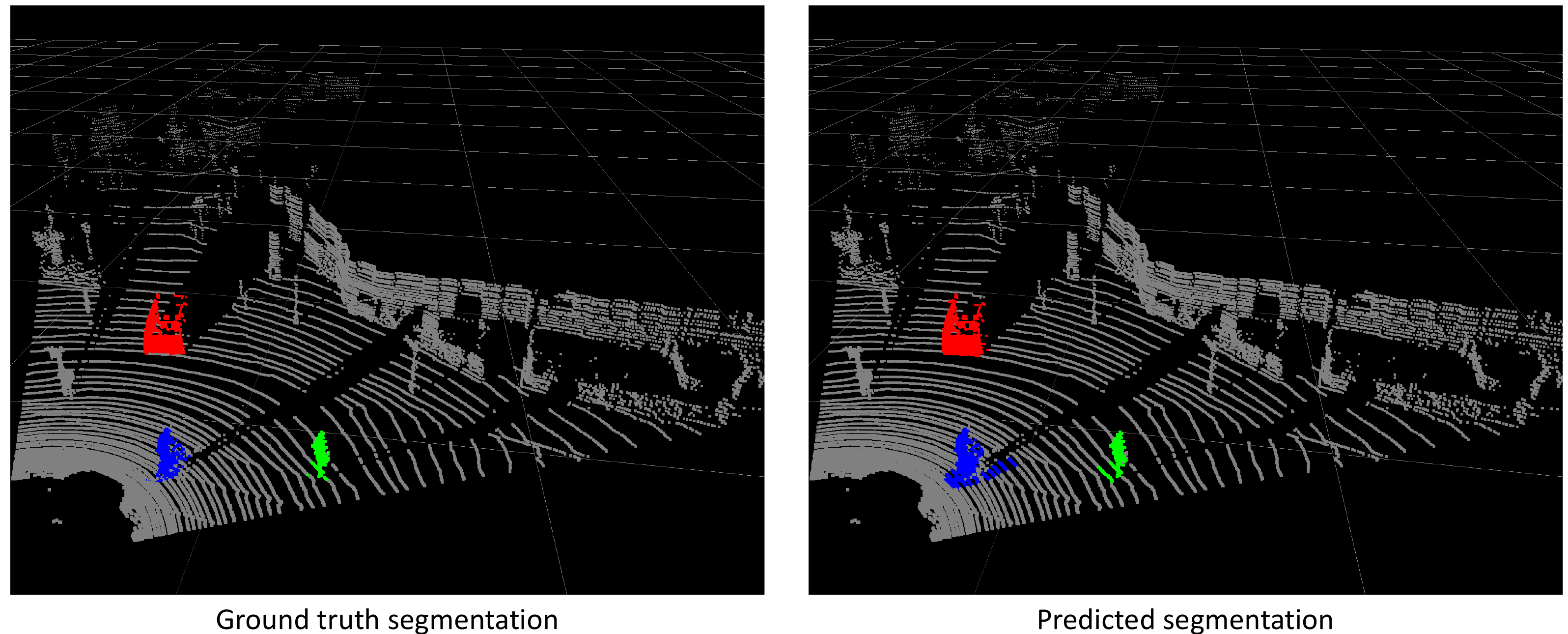}
  \caption{LiDAR point cloud segmentation}
  \label{fig:lidar_seg}
\end{figure}

\sloppy
Our experiments are performed on the task of LiDAR point cloud segmentation; specifically, given a point cloud detected by a LiDAR sensor, we wish to perform point-wise classification, as shown in Fig. \ref{fig:lidar_seg}. This task is an essential step for autonomous vehicles to perceive and understand the environment, and navigate accordingly.

To evaluate the accuracy of the point cloud segmentation algorithm, we compute \textit{Intersection-over-Union} (IoU), \textit{Precision} and \textit{Recall} as:
%between the predicted and the ground truth labels as
\begin{gather*}
IoU_c = \frac{|\mathcal{P}_c \cap \mathcal{G}_c|}{|\mathcal{P}_c \cup \mathcal{G}_c|},
Pr_c = \frac{|\mathcal{P}_c \cap \mathcal{G}_c|}{|\mathcal{P}_c|}, 
Recall_c = \frac{|\mathcal{P}_c \cap \mathcal{G}_c|}{|\mathcal{G}_c|}.
\end{gather*}

Here, $\mathcal{P}_c$ denotes the set of points that our model predicted to be of class-$c$, $\mathcal{G}_c$ denotes the ground-truth set of points belonging to class-$c$, and $|\cdot|$ denotes the cardinality of a set. \textit{Precision} and \textit{Recall} measures accuracy with regard to false positives and false negatives, respectively; while \textit{IoU} takes both into account. For this, \textbf{\textit{IoU}} is used as the primary accuracy metric in our experiments.

\subsection{Experimental Setup}
\label{ssec:experimental_setup}
Our analysis is based on SqueezeSeg \cite{squeezeSeg}, a convolutional neural network based model for point cloud segmentation. To collect the real-word dataset, we used LiDAR point cloud data from the KITTI dataset and converted its 3D bounding box labels to point-wise labels. Since KITTI dataset only provides reliable 3D bounding boxes for front-view LiDAR point clouds, we limit the horizontal field of view(FOV) to the forward-facing $90^\circ$. 
This way, we obtained 10,848 LiDAR scans with manual labels. We used 8,057 scans for training and 2,791 scans for validation. Each point in a KITTI LiDAR scan has 3 cartesian coordinates $(x,y,z)$ and an intensity value, which measures the amplitude of the signal received. Although the intensity measurement as an extra input feature is beneficial to improve the segmentation accuracy, simulating the intensity measurement is very difficult and not supported in our current framework. Therefore we excluded intensity as an input feature to the neural network for GTA-V synthetic data. We use NVIDIA TITAN X GPUs for the experiments during both the training and validation phases.

%and we used our data synthesis framework to generate 8,585 frames as extra training data
\subsection{Experimental Results}

For the first set of experiments, we used our data synthesis framework to generate 8,585 LiDAR point cloud scans in autonomous-driving scenes. The generated data contain $(x, y, z)$ measurements but do not contain intensity. The horizontal FOV of the collected point clouds are set to be $90^\circ$ to match the setting of KITTI point clouds described in Section \ref{ssec:experimental_setup}.

To quantify the effect of training the model with synthetic data, we first trained two models on the KITTI training set with intensity included and excluded, and validated on the KITTI validation set. The performance is shown in the first 2 rows of Table \ref{tab:gta} as the baseline. The model with intensity achived better result. Then we trained another model with only GTA-V synthetic data. As shown in the third row of Table \ref{tab:gta}, the performance drops a lot. This is mostly because the distributions of the synthetic dataset and KITTI dataset are quite different. Therefore, through training purely on synthetic dataset, it is hard for the neural network to learn all the required details for the KITTI dataset, which might be missing or insufficient in the synthetic training dataset. Finally, we combined the KITTI data and GTA-V data together as the training set and train another model. As shown in the last row of Table \ref{tab:gta}, the performance is improved significantly,  almost 9\% better than the accuracy achieved only using real-world data. Despite the loss of the intensity channel, the GTA+KITTI dataset gives better accuracy than if intensity is included. This demonstrates the efficacy of the synthetic data extracted in our framework. 
%To make further comparisons, we trained a fourth model only using KITTI data, but with intensity included as an extra feature (First row in Table \ref{tab:gta}). 

%These experiments also show the effectiveness of the Intensity channel. I

\begin{table}[h]
\centering
\caption{Segmentation Performance Comparison on the Car Category. Only data used in the first row has Intensity channel.}
\label{tab:gta}
\begin{tabular}{cccc}
                                                                                           & Precision & Recall & \textbf{IoU}  \\ \hline\hline
\multicolumn{1}{c|}{KITTI w/ Intensity}                                                    & 66.7      & 95.4   & 64.6          \\
\multicolumn{1}{c|}{KITTI w/o Intensity}                                                   & 58.9      & 95.0   & 57.1          \\
\multicolumn{1}{c|}{GTA-V only}                                                            & 30.4      & 86.6   & 29.0          \\ \hline
\multicolumn{1}{c|}{\begin{tabular}[c]{@{}c@{}}KITTI w/o Intensity\\ + GTA-V\end{tabular}} & 69.6      & 92.8   & \textbf{66.0} \\ \hline
\end{tabular}
\begin{tablenotes}
\vspace{1mm}
\small \item All numbers are in percentage.
\end{tablenotes}
\end{table}

\begin{figure}[h]
  \centering
  \includegraphics[width=3.3in, trim={0cm 0cm 0cm 0cm}]{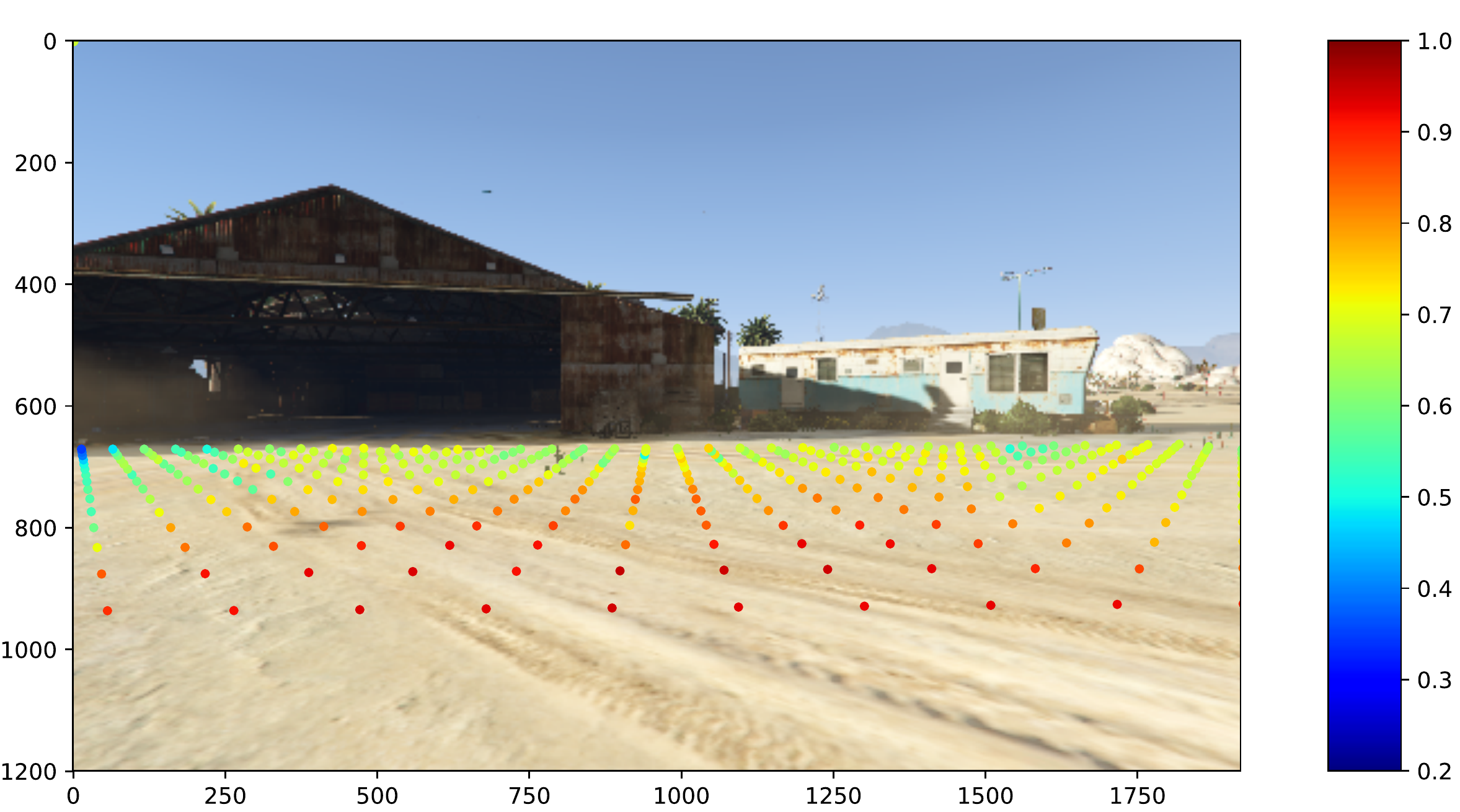}
  \caption{IoU scatter with the change of car location}
  \label{fig:exp_result}
\end{figure}

Then we used our framework to systematically test SqueezeSeg. As an illustrative experiment, we only performed sampling in the car location X-Y dimensions as in Fig. \ref{fig:exp_with_lidar},  rather than the whole modification space. 555 scenes were sampled to test SqueezeSeg, with the IoU results shown in Fig. \ref{fig:exp_result}. The blue and green dots show the car locations resulting in low IoU. Most of the "blind spot" are locations far from the LiDAR scanner, but there are also closer locations that result in low IoU scores. Close locations with low IoUs are dangerous in autonomous driving, since they can mislead the decision-making system of the autonomous vehicles and result in immediate accident.

Further experiments are then done to show the efficacy of using synthetic data from the proposed framework to possibly improve performance of the network on bad sample points in the modification space. We synthesized totally 2,250 LiDAR point cloud scans in 15 different scene backgrounds. In each scene background, only one car is placed with the same orientation as the camera view. We obtained 150 point cloud scans in each scene background by changing the position of the car $(X, Y)$ in the sampled space: $ S=\{(x, y) \mid x \in \{-5,\cdots, 4\}, y \in \{5, \cdots, 19\} \}$, where $X, Y$ are respectively the left-right and forward-backward offset relative to the position of the camera. For each scene background, the position and orientation of the camera were fixed. 

We split the collected point cloud scans based on the scene background. 1200 point cloud scans in the first 7 backgrounds are used as validation set $\mathcal{V}$, and the rest 1050 scans, which we call retraining set $\mathcal{R}$, are used for retraining purpose. First, we train a neural network with purely KITTI data and do evaluation on the synthetic 1200 scans in the validation set. 
We define \textit{mean IoU(mIoU)} for each point in the $15\times10$ X-Y modification space as averaging IoUs over all the 7 scene backgrounds:
\vspace{-1.8mm}
$$mIoU(i,j)=\frac{1}{n}\sum_{k=1}^{n}IoU(i,j,k),$$
where $n$ is the number of scene backgrounds ($n=7$ in this experiment), $(i, j)$ is in $\{(i, j) \mid i \in [-5, 4], j \in [5, 19], i,j \in \mathbb{Z} \}$ and $IoU(i,j,k)$ refers to the IoU of the point cloud scan sampled at $(i,j)$ in the X-Y modification with the $k_{th}$ scene background. 

\begin{figure}[h]
  \centering
  \includegraphics[width=3.8in, trim={3.6cm 2.5cm 2cm 1.cm}]{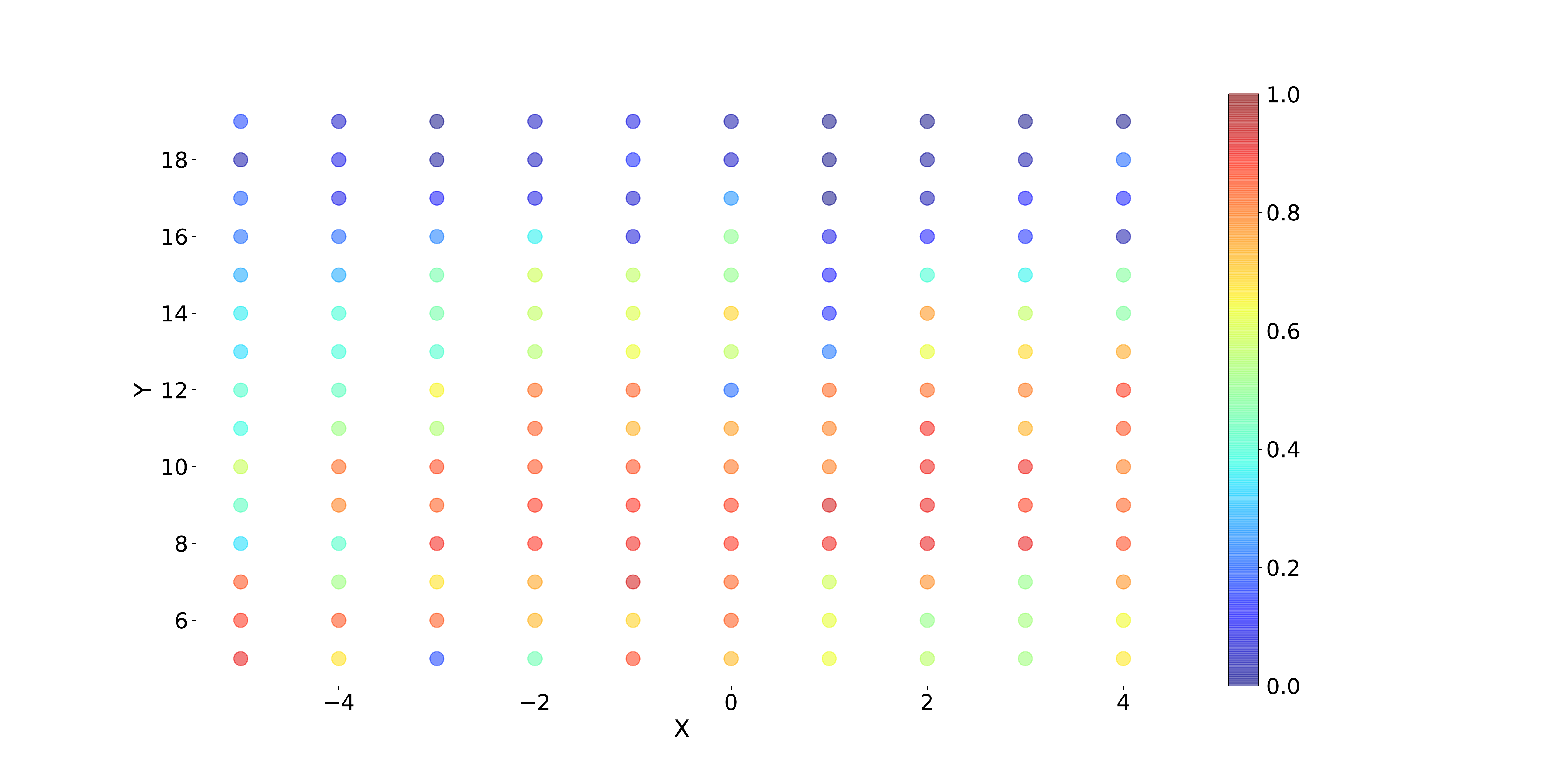}
  %\fbox{\rule[-.5cm]{0cm}{4cm} \rule[-.5cm]{4cm}{0cm}}
  \caption{mIoU map of the validation set before retraining}
  \label{fig:mIoU_before}
\end{figure}

The mIoU map of the validation set is computed, as shown in Fig. \ref{fig:mIoU_before}. We can see that the pre-trained network performs poorly on positions that are far away, at the boundary of the FOV. But more surprisingly, we also observed that on position (-3, 5), which is fairly close to the ego-vehicle, the mIoU score is also very low. Detection errors at such near distance can be very dangerous.

\begin{figure}
  \centering
  \includegraphics[width=3.8in, trim={3.6cm 2.5cm 2cm 1.cm}]{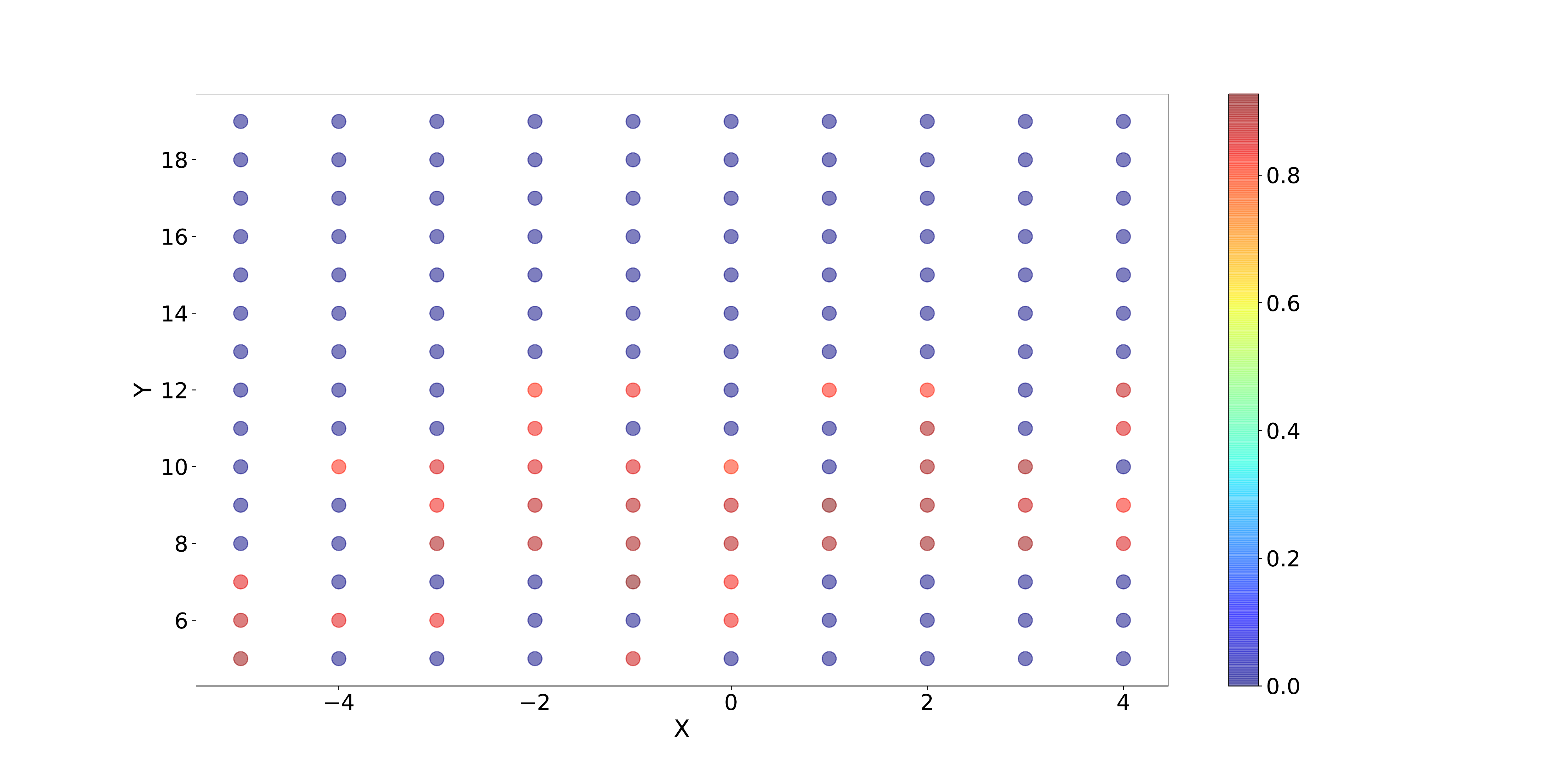}
  %\fbox{\rule[-.5cm]{0cm}{4cm} \rule[-.5cm]{4cm}{0cm}}
  \caption{mIoU map of the validation set after selection with mIoU less than 0.65 set to 0. All the point clouds in the retraining set $\mathcal{R}$ corresponding to the blue positions in the new mIoU map will be added to the original training set.}
  \label{fig:mIoU_selected}
\end{figure}

Based on the mIoU map, we choose positions with an mIoU smaller than a threshod to form a retraining set, as shown in Fig. \ref{fig:mIoU_selected}. 
\begin{figure}
  \centering
  \includegraphics[width=3.8in, trim={3.6cm 2.5cm 2cm 1.cm}]{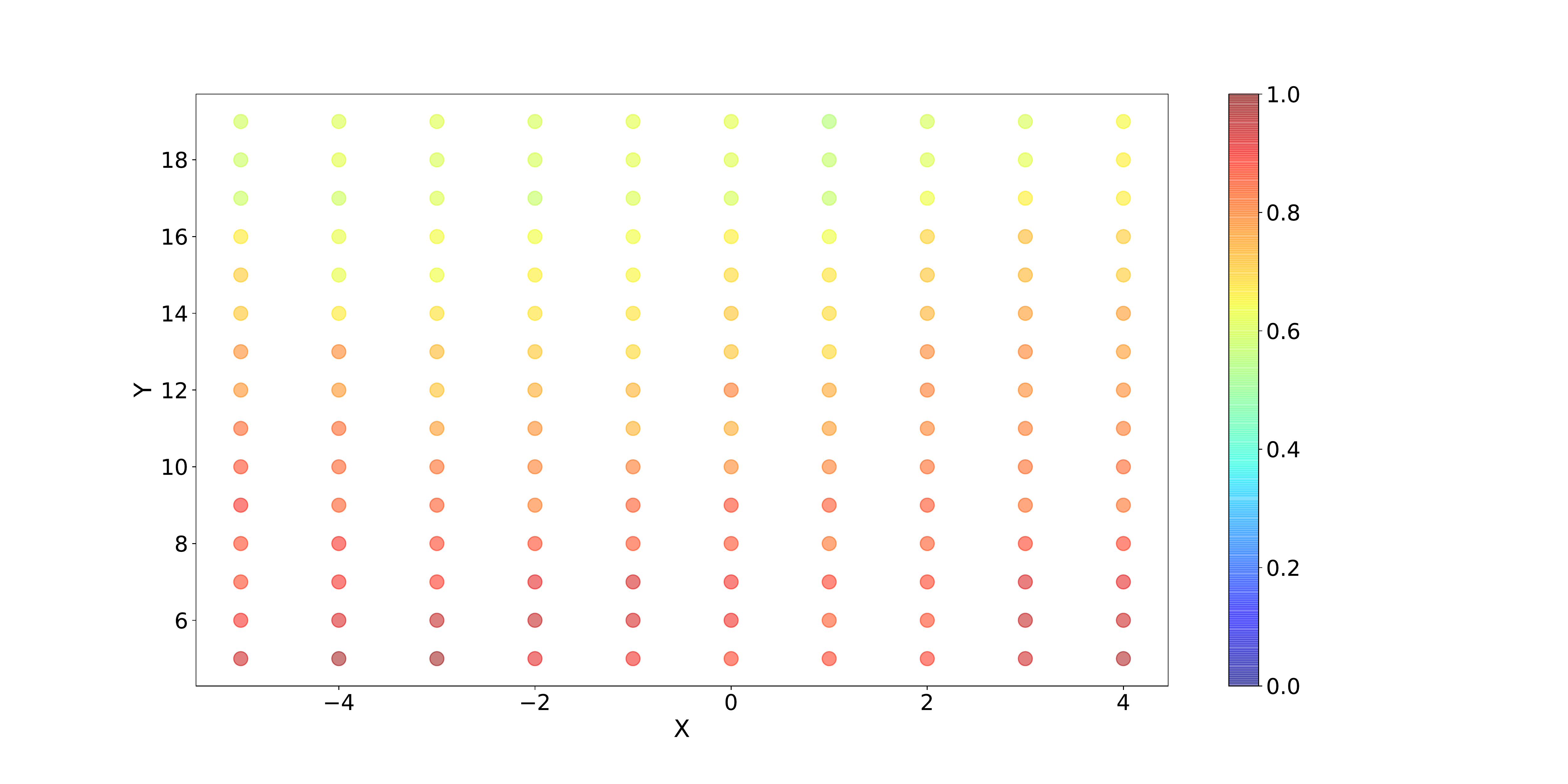}
  %\fbox{\rule[-.5cm]{0cm}{4cm} \rule[-.5cm]{4cm}{0cm}}
  \caption{mIoU map of the validation set after retraining}
  \label{fig:mIoU_retrained}
\end{figure}

\begin{figure}
  \centering
  \includegraphics[width=3.8in, trim={3.6cm 2.5cm 2cm 2.cm}]{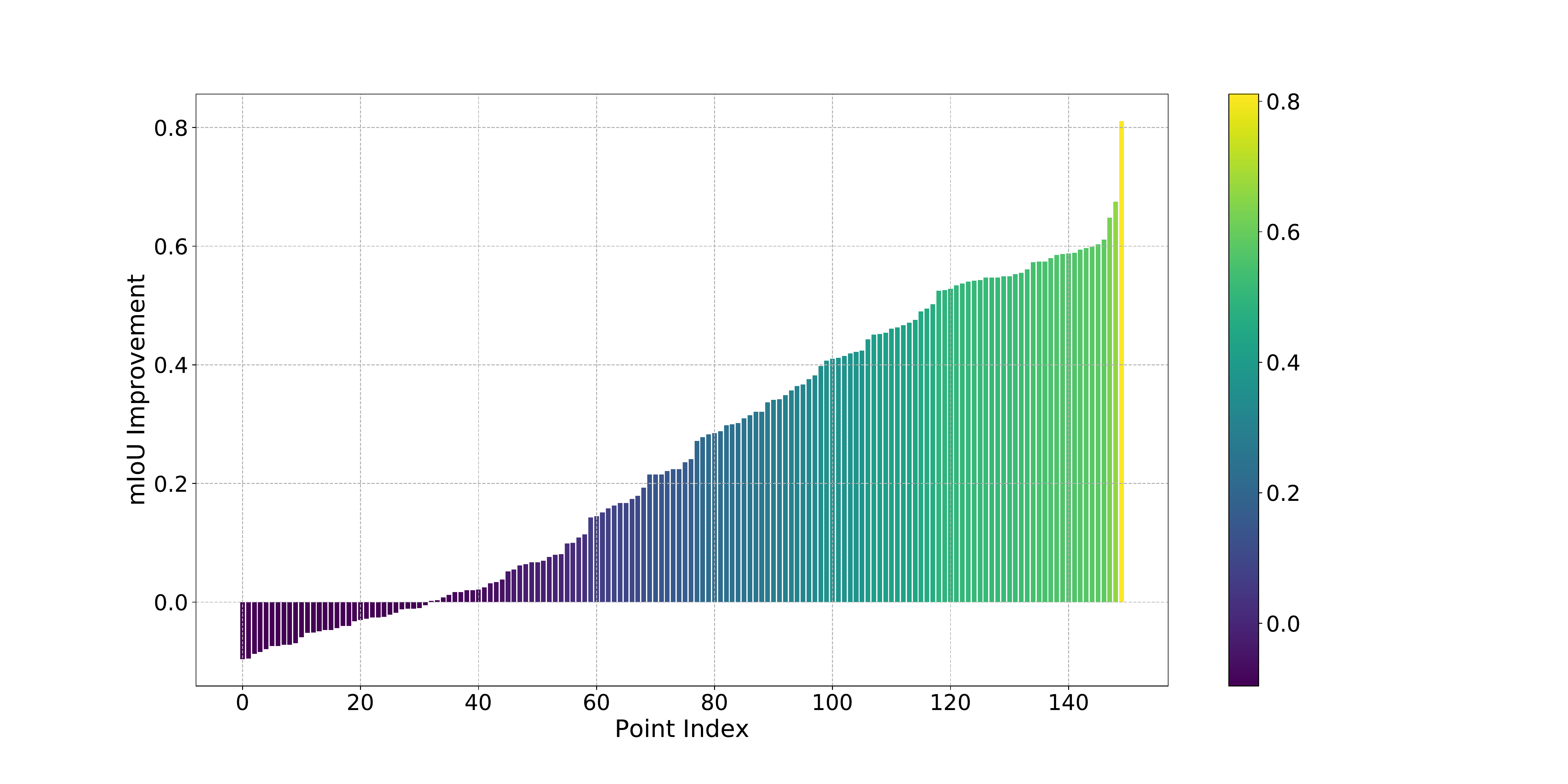}
  %\fbox{\rule[-.5cm]{0cm}{4cm} \rule[-.5cm]{4cm}{0cm}}
  \caption{mIoU improvements in ascending order for all 150 positions}
  \label{fig:mIoU_improvement}
\end{figure}
Then all the point clouds in the retraining set $\mathcal{R}$ with a selected position are added to the original training set. After the retraining process, we re-evaluate the validation set, with the new mIoU map shown in Fig. \ref{fig:mIoU_retrained}. As the figure shows, at almost all the close-to-center positions originally with low mIoU, the neural network performs much better than before the retraining. In order to visualize the performance improvements better, we plot the \textit{mIoU} improvement after the retraining process for each position. The mIoU improvements are sorted and plotted in Fig. \ref{fig:mIoU_improvement}. We see that after retraining, performance on point clouds at most of the positions gets much better, with slightly degraded performance at only a small fraction of positions. Meanwhile, the performance on KITTI dataset remained almost the same with \textit{IoU} changing from 60.8\% to 60.6\%. These experiments show the efficacy of using synthetic data from user-configured scenes of the proposed framework to test, analyze and improve the performance of neural networks through retraining.

%The segmentation accuracy can be potentially boosted by adding more samples near the "blind spot" locations detected by the testing process.

\section{Conclusions and Future Work}
\label{sec:conclusions}
In this paper, we proposed a framework that synthesizes annotated LiDAR point clouds from a virtual world in a game, with a method to automatically calibrate the point cloud and scene image. Our framework can be used to: 1) obtain a large amount of annotated point cloud data, which can then be used to help neural network training; 2) systematically test, analyze and improve performance of neural networks for tasks such as point cloud segmentation. Experiments show that for a point cloud segmentation task, synthesized data help improve the validation accuracy (IoU) by 9\%. Furthermore, the systematical sampling and testing framework can help us to identify potential weakness/blind spots of our neural network model and fix them. The first set of experiments also show the effectiveness of the intensity channel in LiDAR point clouds. In the future, we will work on simulating the intensity information, which we believe will definitely help the research in this field.

{\small
  \bibliography{bibliography.bib}

% Generated by IEEEtran.bst, version: 1.14 (2015/08/26)
\begin{thebibliography}{10}
\providecommand{\url}[1]{#1}
\csname url@samestyle\endcsname
\providecommand{\newblock}{\relax}
\providecommand{\bibinfo}[2]{#2}
\providecommand{\BIBentrySTDinterwordspacing}{\spaceskip=0pt\relax}
\providecommand{\BIBentryALTinterwordstretchfactor}{4}
\providecommand{\BIBentryALTinterwordspacing}{\spaceskip=\fontdimen2\font plus
\BIBentryALTinterwordstretchfactor\fontdimen3\font minus
  \fontdimen4\font\relax}
\providecommand{\BIBforeignlanguage}[2]{{%
\expandafter\ifx\csname l@#1\endcsname\relax
\typeout{** WARNING: IEEEtran.bst: No hyphenation pattern has been}%
\typeout{** loaded for the language `#1'. Using the pattern for}%
\typeout{** the default language instead.}%
\else
\language=\csname l@#1\endcsname
\fi
#2}}
\providecommand{\BIBdecl}{\relax}
\BIBdecl

\bibitem{3dlidar}
F.~Moosmann, O.~Pink, and C.~Stiller, ``Segmentation of 3d lidar data in
  non-flat urban environments using a local convexity criterion,'' in
  \emph{IEEE Intelligent Vehicles Symposium}, 2009, pp. 215--220.

\bibitem{squeezeSeg}
B.~Wu, A.~Wan, X.~Yue, and K.~Keutzer, ``Squeezeseg: Convolutional neural nets
  with recurrent crf for real-time road-object segmentation from 3d lidar point
  cloud,'' in \emph{IEEE International Conference on Robotics and Automation},
  2018.

\bibitem{segmentation1}
A.~Dewan, G.~L. Oliveira, and W.~Burgard, ``Deep semantic classification for 3d
  lidar data,'' \emph{Computing Research Repository}, vol. abs/1706.08355,
  2017.

\bibitem{segmentation2}
D.~Dohan, B.~Matejek, and T.~Funkhouser, ``Learning hierarchical semantic
  segmentations of {LIDAR} data,'' in \emph{International Conference on 3D
  Vision (3DV)}, Oct. 2015.

\bibitem{drivable1}
Z.~Liu, S.~Yu, X.~Wang, and N.~Zheng, ``Detecting drivable area for
  self-driving cars: An unsupervised approach,'' \emph{arXiv preprint
  arXiv:1705.00451}, 2017.

\bibitem{drivable2}
R.~Fernandes, C.~Premebida, P.~Peixoto, D.~Wolf, and U.~Nunes, ``Road detection
  using high resolution lidar,'' in \emph{IEEE Vehicle Power and Propulsion
  Conference (VPPC)}, Oct 2014, pp. 1--6.

\bibitem{indoorRGBD}
N.~Silberman, D.~Hoiem, P.~Kohli, and R.~Fergus, ``Indoor segmentation and
  support inference from rgbd images,'' in \emph{European Conference on
  Computer Vision (ECCV)}, 2012, pp. 746--760.

\bibitem{KITTI}
A.~Geiger, P.~Lenz, C.~Stiller, and R.~Urtasun, ``Vision meets robotics: The
  kitti dataset,'' \emph{International Journal of Robotics Research (IJRR)},
  2013.

\bibitem{rapid3DSelection}
R.~Kopper, F.~Bacim, and D.~A. Bowman, ``Rapid and accurate 3d selection by
  progressive refinement,'' in \emph{IEEE Symposium on 3D User Interfaces
  (3DUI)}, March 2011, pp. 67--74.

\bibitem{goThenTag}
M.~Veit and A.~Capobianco, ``Go'then'tag: A 3-d point cloud annotation
  technique,'' in \emph{IEEE Symposium on 3D User Interfaces (3DUI)}.\hskip 1em
  plus 0.5em minus 0.4em\relax IEEE, 2014, pp. 193--194.

\bibitem{smartAnnotator}
Y.-S. Wong, H.-K. Chu, and N.~J. Mitra, ``Smartannotator: An interactive tool
  for annotating {RGBD} indoor images,'' \emph{arXiv preprint arXiv:1403.5718},
  2014.

\bibitem{interactiveRGBD}
T.~Shao, W.~Xu, K.~Zhou, J.~Wang, D.~Li, and B.~Guo, ``An interactive approach
  to semantic modeling of indoor scenes with an rgbd camera,'' \emph{ACM Trans.
  Graph.}, vol.~31, no.~6, pp. 136:1--136:11, 2012.

\bibitem{groupAnnotation}
A.~Boyko and T.~Funkhouser, ``Cheaper by the dozen: Group annotation of 3d
  data,'' in \emph{Proceedings of the 27th Annual ACM Symposium on User
  Interface Software and Technology}, ser. UIST '14.\hskip 1em plus 0.5em minus
  0.4em\relax ACM, 2014, pp. 33--42.

\bibitem{activelearning1}
A.~Kapoor, K.~Grauman, R.~Urtasun, and T.~Darrell, ``Active learning with
  gaussian processes for object categorization,'' in \emph{IEEE 11th
  International Conference on Computer Vision (ICCV)}, 2007, pp. 1--8.

\bibitem{activelearning2}
A.~Top, G.~Hamarneh, and R.~Abugharbieh, ``Active learning for interactive 3d
  image segmentation,'' in \emph{International Conference on Medical Image
  Computing and Computer-Assisted Intervention}, 2011, pp. 603--610.

\bibitem{onlineCrowdsourcing}
P.~Welinder and P.~Perona, ``Online crowdsourcing: rating annotators and
  obtaining cost-effective labels,'' in \emph{IEEE Computer Society Conference
  on Computer Vision and Pattern Recognition Workshops (CVPRW),}.\hskip 1em
  plus 0.5em minus 0.4em\relax IEEE, 2010, pp. 25--32.

\bibitem{playingForData}
S.~R. Richter, V.~Vineet, S.~Roth, and V.~Koltun, ``Playing for data: Ground
  truth from computer games,'' in \emph{European Conference on Computer Vision
  (ECCV)}, 2016, pp. 102--118.

\bibitem{drivingInMatrix}
M.~Johnson-Roberson, C.~Barto, R.~Mehta, S.~N. Sridhar, K.~Rosaen, and
  R.~Vasudevan, ``Driving in the matrix: Can virtual worlds replace
  human-generated annotations for real world tasks?'' in \emph{Robotics and
  Automation (ICRA), 2017 IEEE International Conference on}.\hskip 1em plus
  0.5em minus 0.4em\relax IEEE, 2017, pp. 746--753.

\bibitem{playing_for_benchmarks}
S.~R. Richter, Z.~Hayder, and V.~Koltun, ``Playing for benchmarks,'' in
  \emph{International Conference on Computer Vision (ICCV)}, 2017.

\bibitem{congrats}
D.~Biedermann, M.~Ochs, and R.~Mester, ``Evaluating visual adas components on
  the congrats dataset,'' in \emph{IEEE Intelligent Vehicles Symposium (IV)},
  June 2016, pp. 986--991.

\bibitem{simulator}
V.~Haltakov, C.~Unger, and S.~Ilic, ``Framework for generation of synthetic
  ground truth data for driver assistance applications,'' in \emph{German
  Conference on Pattern Recognition}.\hskip 1em plus 0.5em minus 0.4em\relax
  Springer, 2013, pp. 323--332.

\bibitem{carla}
A.~Dosovitskiy, G.~Ros, F.~Codevilla, A.~Lopez, and V.~Koltun, ``{CARLA}: {An}
  open urban driving simulator,'' in \emph{Proceedings of the 1st Annual
  Conference on Robot Learning}, 2017, pp. 1--16.

\bibitem{dreossiDS17}
T.~Dreossi, A.~Donz{\'{e}}, and S.~A. Seshia, ``Compositional falsification of
  cyber-physical systems with machine learning components,'' \emph{Computing
  Research Repository}, vol. abs/1703.00978, 2017.

\bibitem{falsifyCNN}
T.~Dreossi, S.~Ghosh, A.~Sangiovanni-Vincentelli, and S.~A. Seshia,
  ``Systematic testing of convolutional neural networks for autonomous
  driving,'' \emph{arXiv preprint arXiv:1708.03309}, 2017.

\end{thebibliography}
  \bibliographystyle{IEEEtran}
}

\end{document}